\documentclass[runningheads]{llncs}
\usepackage{graphicx}
\usepackage{caption,subcaption}
\usepackage{multirow}
\usepackage{color, colortbl, xcolor}
\usepackage[title]{appendix}

\usepackage{mathtools}
\usepackage{amssymb}
\usepackage{float}
\usepackage{booktabs}
\usepackage[colorlinks=true, urlcolor=blue, citecolor={blue!50!black}, pdfborder={0 0 0}]{hyperref}
\usepackage{tcolorbox}

\definecolor{cerulean}{rgb}{0.0, 0.48, 0.65}
\definecolor{copper}{rgb}{0.72, 0.45, 0.2}
\definecolor{cadmiumgreen}{rgb}{0.0, 0.42, 0.24}

\begin{document}
\title{Fusion Models for Improved Visual Captioning}

\author{Marimuthu Kalimuthu\thanks{corresponding author} \and
Aditya Mogadala \and
\small{Marius Mosbach}  \and
\small{Dietrich Klakow}
}
\authorrunning{M. Kalimuthu et al.}
\institute{Spoken Language Systems (LSV) \\ Saarland Informatics Campus, Saarland University \\ Saarbr\"ucken, Germany \\
\email{\{mkalimuthu,amogadala,mmosbach,dklakow\}@lsv.uni-saarland.de}}
\maketitle              % typeset the header of the contribution
\begin{abstract}

Visual captioning aims to generate textual descriptions given images or videos. Traditionally, image captioning models are trained on human annotated datasets such as Flickr30k and MS-COCO, which are limited in size and diversity. This limitation hinders the generalization capabilities of these models while also rendering them liable to making mistakes. Language models can, however, be trained on vast amounts of freely available unlabelled data and have recently emerged as successful language encoders~\cite{bert-devlin:2019} and coherent text generators~\cite{gpt-3-brown:2020}. Meanwhile, several unimodal and multimodal fusion techniques have been proven to work well for natural language generation~\cite{hnsg-fan:2018} and automatic speech recognition~\cite{coldfusion-sriram:2018}. Building on these recent developments, and with the aim of improving the quality of generated captions, the contribution of our work in this paper is two-fold: First, we propose a generic multimodal model fusion framework for caption generation as well as emendation where we utilize different fusion strategies to integrate a pretrained Auxiliary Language Model (AuxLM) within the traditional encoder-decoder visual captioning frameworks. Next, we employ the same fusion strategies to integrate a pretrained Masked Language Model (MLM), namely BERT, with a visual captioning model, viz. \textit{Show, Attend, and Tell}, for emending both syntactic and semantic errors in captions. Our caption emendation experiments on three benchmark image captioning datasets, viz. Flickr8k, Flickr30k, and MSCOCO, show improvements over the baseline, indicating the usefulness of our proposed multimodal fusion strategies. Further, we perform a preliminary qualitative analysis on the emended captions and identify error categories based on the type of corrections.


\end{abstract}
\section{Introduction}
\label{sec:intro}
The field of deep learning has seen tremendous progress ever since the breakthrough results of AlexNet~\cite{alexnet-kriz:2012} on the ImageNet Large Scale Visual Recognition Challenge (ILSVRC). Significant algorithmic improvements have since been achieved, both for language~\cite{bert-devlin:2019,attention-vaswani:2017} and visual scene understanding~\cite{detr-carion:2020,resnet-he:2016}. Further, there has been numerous efforts for the joint understanding of language and visual modalities in terms of defining novel tasks, creating benchmark datasets, and proposing new methods to tackle the challenges arising out of the multimodal nature of data~\cite{vision-and-language-mogadala:2020}. One specific problem that has piqued the interest of researchers and practitioners in Computer Vision (CV) and Natural Language Processing (NLP) is \textit{Visual Captioning}. Considerable progress has been made since the introduction of \textit{Show and Tell}~\cite{show-and-tell-vinyals:2015}, which is an end-to-end model for image captioning. 
Despite these encouraging developments, the visual captioning models are still brittle, unreliable, prone to making unexplainable mistakes~\cite{hallucination-rohrbach:2018}, and, most importantly, their performance is nowhere close to human-level understanding. However, although still a black-box, the recent unprecedented progress on language modeling has proven the potential of these models to generate semantically fluent and coherent text~\cite{gpt-3-brown:2020,gpt-2-radford:2019} and encode powerful representations of language using bidirectional context~\cite{bert-devlin:2019}. Building on these developments, we propose to incorporate external language models into visual captioning frameworks to aid and improve their capabilities both for description generation and emendation. Although the proposed architecture~(See Figure~\ref{fig:fusion-arch}) could be used for both caption generation and correction, in this paper, we only focus on the task of emending captions. However, we describe the changes needed to the architecture to achieve caption generation. Broadly, our architecture consists of four major components, viz. a \textit{CNN encoder}, an \textit{LSTM decoder}, a \textit{pretrained AuxLM}, and a \textit{Fusion module} (See Figure~\ref{fig:fusion-arch}). To the best of our knowledge, our architecture is novel since the current caption editing approaches~\cite{look-modify-sammani:2019,edit-tell-sammani:2020} in the literature do not leverage AuxLMs.

The rest of the paper is organized as follows. In Section~\ref{sec:rel-work}, we briefly review relevant works on caption editing. In Section~\ref{sec:fusion-tech}, we provide background on fusion strategies that have been empirically shown to work well for neural machine translation, automatic speech recognition, and story generation and introduce our fusion architecture. Following this, we describe our experimental setup including the implementation details of our caption editing model in Section~\ref{sec:experiments}. We then present our quantitative results along with a short qualitative analysis of the emended captions in Section~\ref{sec:results}. Finally, we conclude our paper in Section~\ref{sec:conc} discussing some future research directions.

\section{Related Work}
\label{sec:rel-work}
Recently, few approaches have been proposed for editing image captions~\cite{look-modify-sammani:2019,edit-tell-sammani:2020}. Although these methods have been shown to produce improved quantitative results, they have some limitations such as completely relying on \textit{labelled} data, which is limited in size and diversity, for training. In addition, it is cost-intensive to obtain caption annotations and, sometimes, it is not even possible due to privacy concerns.  This is the case, for instance, in the medical domain. Furthermore, these approaches implicitly learn a language model on the decoder-side with limited textual data. Moreover, they do not make use of external language models, i.e., AuxLMs, which can be trained on freely available unlabelled data, are more powerful since they learn rich language representations~\cite{bert-devlin:2019,gpt-2-radford:2019}, and recently have been trained on enormous amounts of data with billions of parameters~\cite{gpt-3-brown:2020,gpt-2-radford:2019}. To the best of our knowledge, there have been no previous works that incorporate AuxLMs into the caption model either for description generation or to correct errors. To address the above stated limitations and leverage the advantages of AuxLMs as powerful language learners, we propose a generic framework  that can accomplish both caption generation and editing tasks depending on the type of AuxLM used.

\section{Fusion Techniques and Variations}
\label{sec:fusion-tech}

Unimodal and multimodal model fusion has been explored extensively in the context of ASR~\cite{memory-control-fusion-cho:2019,coldfusion-sriram:2018}, Neural Machine Translation (NMT)~\cite{shallow-fusion-gulcehre:2015}, and hierarchical story generation~\cite{hnsg-fan:2018}. However, to the best of our knowledge, there have been no similar works for visual captioning. In this section, we review these fusion methods and relate them to our goal of achieving image caption generation and emendation by using additional language models, i.e., AuxLMs.

\subsubsection{Deep Fusion.}
Gulcehre et al.~\cite{shallow-fusion-gulcehre:2015} explored \textit{Deep Fusion} for NMT in which a translation model trained on parallel corpora and an AuxLM trained on monolingual, target language data are combined using a trainable single layer neural network model. Typically, the AuxLM is an autoregressive LSTM-based recurrent neural network~\cite{rnnlm-mikolov:2010} while the translation model follows a typical sequence-to-sequence (Seq2Seq)~\cite{seq2seq-cho:2014} encoder-decoder architecture. To achieve \textit{Deep Fusion}, a gate is learned from the hidden state of AuxLM and then concatenated with the hidden state of the translation model.

The drawback of this approach is that both the AuxLM and the translation model are trained separately and kept frozen while the fusion layer is trained. Hence, they never get a chance to communicate and adapt their parameters \textit{during} training. This forces these models to learn redundant representations.

\subsubsection{Cold Fusion.}
To address the limitation of \textit{Deep Fusion}, Sriram et al.~\cite{coldfusion-sriram:2018} introduced the so-called \textit{Cold Fusion} strategy for ASR. The mechanism uses a pretrained AuxLM while training the Seq2Seq model. The AuxLM is still kept frozen as in the case of \textit{Deep Fusion}, however the parameters of the fusion layer and the Seq2Seq decoder are trained together, leveraging the already-learned language representations of AuxLM. For more details see Section~\ref{ssec:cold-fusion}.

\subsubsection{Hierarchical Fusion.}
In a similar vein, Fan et al.~\cite{hnsg-fan:2018} has explored a sophisticated fusion strategy for textual story generation in a hierarchical fashion where they combine a \textit{pretrained} convolutional Seq2Seq model with a trainable convolutional Seq2Seq model. More specifically, the hidden states of a pretrained Seq2Seq model are fused with the hidden states of a trainable Seq2Seq model using a slightly modified version of the \textit{Cold Fusion}~\cite{coldfusion-sriram:2018} approach. We call this scheme as \textit{Hierarchical Fusion} hereafter.

Considering the advantages of the fusion techniques of Sriram et al.~\cite{coldfusion-sriram:2018} and Fan et al.~\cite{hnsg-fan:2018} over \textit{Deep Fusion}, we adapt them in our fusion model framework.

\subsection{Auxiliary Language Model}
\label{ssec:bert-auxlm}
Depending on the task, i.e., caption generation or editing, two classes of language models can be used as AuxLMs. For caption generation, since we will not have access to the right-side context during test time, an autoregressive (i.e., causal) language model~\cite{neural-problm-bengio:2003,gpt-3-brown:2020,lstm-lm-merity:2018,gpt-2-radford:2019} that predicts the next token using only the left-side context is qualified to be used as the AuxLM. For a caption editing task, however, a bidirectionally contextualized language model, for example BERT~\cite{bert-devlin:2019}, would be better suited as AuxLM since they better encode sequences than their unidirectional counterparts. Also, intuitively, since our task is to edit captions, we already have a noisy version of captions (either from the baseline or from other off-the-shelf caption models). If any errors are detected, we want to correct them, but otherwise leave them unmodified. In our experiments, we use a pretrained BERT model (uncased) that has been trained on English Wikipedia and the BooksCorpus using a combination of Masked Language Modeling (MLM) and next-sentence prediction objectives. Although it is possible to fine-tune the pretrained AuxLMs~\cite{bert-finetune-mosbach:2020} on the target domain (the image captions in our case) before integrating it in the fusion model, we did not perform this step in our experiments. However, the fine tuning process would likely prove to be beneficial if we intend to adapt the AuxLMs to more focused domains such as medical~\cite{imageclefmedcaption-overview-pelka:2020} or remote sensing~\cite{remote-sensing-lu:2017} and we leave this pursuit for future work.

\subsection{Fusion Strategies and Architecture}
\label{ssec:auxlm-fusion}
Inspired by the success of multimodal fusion strategies for ASR~\cite{memory-control-fusion-cho:2019,coldfusion-sriram:2018} and unimodal fusion for NLG~\cite{hnsg-fan:2018,shallow-fusion-gulcehre:2015}, we slightly modify these fusion schemes and utilize them for the task of emending image captions. By doing so, we leverage the rich language representations of AuxLMs to achieve sentence-level coherency and grammatical correctness in the emended captions.

\begin{figure*}[!ht]
    \centering
    \includegraphics[width=\textwidth]{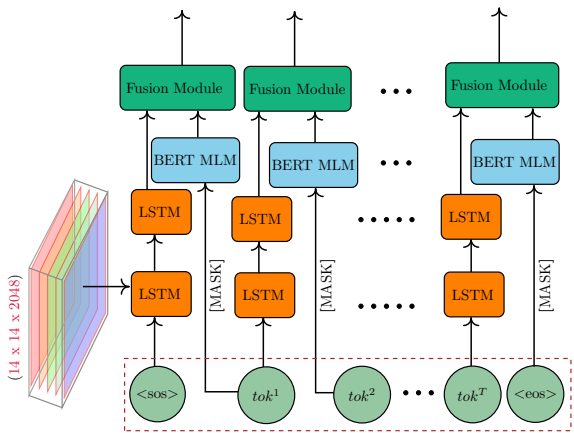}
    \vspace*{-26em}
    \caption{Architecture of our proposed fusion model. The encoded image is fed to the LSTM decoder only at the first time step. \textit{BERT MLM}~\cite{bert-devlin:2019} has been pretrained to predict the masked token at current time step whereas the LSTM decoder is trained to predict the next token given all previous tokens. The \textit{Fusion Module} can be any instance of the fusion schemes discussed in Section~\ref{ssec:auxlm-fusion}.}
  \label{fig:fusion-arch}
\end{figure*}

Figure~\ref{fig:fusion-arch} depicts the \textit{general} architecture of the proposed fusion model. It contains four major components, namely a \textit{ConvNet encoder}, an \textit{LSTM decoder}, a \textit{BERT encoder} (pretrained MLM), and a \textit{Fusion module}. The architecture is flexible and can be used for two tasks, i.e., visual captioning and caption emendation, depending on the type of AuxLM chosen. For caption emendation, a text encoder such as BERT is used as an AuxLM. The LSTM decoder processes the sequence from left-to-right whereas the BERT model utilizes the entire sequence due to its inherent nature of encoding contextualized representation.

For the visual captioning task, the AuxLM must be an autoregressive model since we do not have access to the whole sequence at inference time. One possibility could be to replace the \textit{BERT MLM} component with an LSTM-based AuxLM or the recently proposed (and more powerful) Transformer-based~\cite{attention-vaswani:2017} autoregressive language model GPT-3~\cite{gpt-3-brown:2020}.

Further, for both captioning and emendation tasks, the \textit{Fusion Module} component is flexible enough to support any sophisticated fusion method, which provides a framework and an opportunity for future works to come up with improved fusion schemes. In addition, the architecture could be employed in a domain-adapted way for visual captioning by integrating an AuxLM that has been trained in the domain of interest. There has been growing interest in recent years to automatically generate descriptions for medical images\footnote{\url{https://www.imageclef.org/2020/medical/caption}\label{fnote:imageclef-medcaption-url}} such as radiology outputs~\cite{imageclefmedcaption-kalimuthu:2020,imageclefmedcaption-overview-pelka:2020}. The domain adapted AuxLM can be useful particularly in settings where labelled image caption data is scarce. However, unlabelled textual data is usually abundant. In such scenarios, an AuxLM can be trained on the target domain data and integrated into our fusion model framework for generating target domain specific image descriptions.

We now introduce the notations of hidden states used in our fusion model. Following Sriram et al. \cite{coldfusion-sriram:2018}, we represent the final layer hidden states of pretrained BERT and trainable LSTM decoder as $h^{MLM}$ and $h^{LSTM}$ respectively.

\subsubsection{Simple Fusion (SF).}
\label{ssec:simple-fusion}
One of the simplest possible fusion mechanisms is through the concatenation of the hidden states of pretrained AuxLM and trainable visual captioning model, followed by a single projection layer with some non-linear activation function~($\sigma$), such as ReLU.
\begin{subequations}
\label{eqn:simple-fusion}
\begin{align}
    \textbf{g}_t &= \sigma\left(\textbf{W}[\textbf{h}_t^{LSTM};\textbf{h}_t^{MLM}] + \textbf{b}\right)
\end{align}
\end{subequations}

\noindent The output of above non-linear transformation ($\textbf{g}_t$) can then be passed through a single linear layer with dropout to obtain prediction scores over the vocabulary.

\subsubsection{Cold Fusion (CF).}
\label{ssec:cold-fusion}
A more sophisticated fusion can be achieved by introducing gates, thereby allowing the captioning model and AuxLM to moderate the information flow between them during the training phase. We slightly modify the cold fusion approach of Sriram et al. \cite{coldfusion-sriram:2018} in our fusion model which is as follows:
\begin{subequations}
\label{eqn:cold-fusion}
\begin{align}
    \textbf{h}_t^{LM} &= \sigma\left(\textbf{W}[\textbf{h}_t^{MLM}] + \textbf{b}\right) \\
    \textbf{g}_t &= \sigma\left(\textbf{W}[\textbf{h}_t^{LSTM}; \textbf{h}_t^{LM}] + \textbf{b}\right) \\
    \textbf{h}_t^{CF} &= [\textbf{h}_t^{LSTM};\left(\textbf{g}_t \circ \textbf{h}_t^{LM}\right)] \\
    \textbf{r}_t^{CF} &= \sigma\left(\textbf{W}[\textbf{h}_t^{CF}] + \textbf{b}\right)
\end{align}
\end{subequations}

\noindent As with simple fusion, the representation $\textbf{r}_t^{CF}$ is followed by a single linear layer with dropout (not shown here) to obtain prediction scores over the vocabulary.

\subsubsection{Hierarchical Fusion (HF).}
\label{ssec:hnsg-fusion}
In the context of text generation, an advanced fusion mechanism based on \textit{Cold Fusion} has been introduced by Fan et al. \cite{hnsg-fan:2018} for the open-ended and creative task of story generation. We adopt their way of model fusion with minor modifications, in the spirit of keeping the model simple. More specifically, after learning two separate gates followed by a concatenation, we only use a single linear layer with GLU activations~\cite{glu-dauphin:2017} instead of 5. Further, to capture the rich sequence representation for caption editing, we use an MLM as AuxLM instead of a convolutional Seq2Seq model. We refer to Fan et al.~\cite{hnsg-fan:2018} for full details of their fusion mechanism.
\begin{subequations}
\label{eqn:hnsg-fusion}
\begin{align}
    \textbf{h}_t^{C} &= \left[\textbf{h}_t^{MLM} ; \textbf{h}_t^{LSTM}\right] \\
    \textbf{g}_t^{left} &= \sigma\left(\textbf{W}[\textbf{h}_t^{C}] + \textbf{b}\right) \circ \textbf{h}_t^{C} \\
    \textbf{g}_t^{right} &= \textbf{h}_t^{C} \circ \left(\sigma\left(\textbf{W}[\textbf{h}_t^{C}] + \textbf{b}\right) \right) \\
    \textbf{g}_t^{C} &= {GLU}\left([\textbf{g}_t^{left} ; \textbf{g}_t^{right}]\right) \\
    \textbf{g}_t^{lp} &= \textbf{W}[\textbf{g}_t^{C}] + \textbf{b} \\
    \textbf{g}_t^{f} &= {GLU}\left( \textbf{g}_t^{lp} \right)
\end{align}
\end{subequations}

\noindent Again, the result of the final GLU (i.e., $\textbf{g}_t^{f}$) is passed through a single linear layer with dropout to obtain prediction scores over the image caption vocabulary. \\

\noindent In all our fusion methods (i.e., SF, CF, and HF), \textbf{;} represents concatenation, $\circ$ stands for hadamard product, and $\sigma$ indicates a non-linear activation function, for which we use ReLU. The gating parameters $\textbf{W}$ and $\textbf{b}$, which are part of the \textit{Fusion Module}~(see Figure~\ref{fig:fusion-arch}), are learned while training the LSTM decoder of the fusion model whereas all the parameters of BERT MLM are kept frozen.

\section{Experiments}
\label{sec:experiments}
We train one baseline and three fusion models on the three commonly used image captioning datasets: Flickr8k, Flickr30k, and MS-COCO. Descriptions and implementation details are given in the following sections.

\subsection{Baseline}
\label{ssec:baseline-model}
This will be the model without any AuxLM component in its architecture. Any off-the-shelf visual captioning model satisfying this condition can be used as a baseline. The only requirement is that it should be possible to generate captions given the test set images from the dataset in question. In our experiments, we use \textit{Show, Attend, and Tell}~\cite{show-attend-tell-xu:2015} where we replace the original VGGNet with ImageNet pretrained ResNet-101~\cite{resnet-he:2016} which encodes the images to a feature map of size 14~x~14~x~2048. For the decoder, we use the standard LSTM with two hidden layers. After training, we use a beam size of 5 for generating captions on the test sets of respective datasets.

\subsection{Fusion Model Training}
\label{ssec:fusion-train}
For each of the datasets, we train three caption models with different initializations for all the fusion techniques (i.e., SF, CF, and HF) proposed in Section~\ref{ssec:auxlm-fusion}.
\subsubsection{Implementation Details.}
First, we lowercase the captions and tokenize them using WordPiece\footnote{\url{https://github.com/google-research/bert}\label{fnote:goog-bert-wordpiece-github}} tokenization~\cite{wordpiece-token-wu:2016} in the same way the BERT MLM model was trained. This consistency in tokenization is important for successful training since the captioning model relies on AuxLM for the hidden state representations at all time steps throughout the training and testing phases. Tokens appearing less than 5 times are replaced with a special $<$unk$>$ token yielding a vocabulary size of 25k. We implement our fusion models in PyTorch~\cite{pytorch-paszke:2019}.

\subsubsection{Decoder and Fusion Module Training.} The images are rescaled to a fixed size of 256~x~256 and encoded using ResNet101~\cite{resnet-he:2016} pretrained on ImageNet and kept frozen throughout training. As with the baseline model, we use an LSTM with 2 hidden layers and set the embedding and decoder dimensions to 1024. The LSTM decoder takes the token at current time step along with previous history and predicts the next token. The BERT MLM model, however, consumes the entire sequence with the token at the next time step being masked using a special [MASK] token and it predicts this masked token (See Figure~\ref{fig:fusion-arch}). The hidden state representations of both LSTM decoder and BERT are then passed to the \textit{Fusion Module}, which can be any one of the fusion mechanisms discussed in Section~\ref{ssec:auxlm-fusion}, to predict the next token (seen from the perspective of the LSTM decoder).

We minimize the Cross Entropy loss using the Adam optimizer~\cite{adam-kingma:2015} with a learning rate of $5e^{-4}$ and a batch size of 128. Initially, the model is scheduled to be trained for 7 epochs. However, the learning rate is halved or an early stopping is triggered if the validation BLEU did not improve for 2 and 4 consecutive epochs respectively.

\subsubsection{Caption Emendation.}
After the fusion model is trained, it can be used in inference mode to correct errors in the captions. While decoding, at time-step \textit{t}, the LSTM predicts $\textit{(t+1)}^{th}$ token based on left-side history whereas BERT MLM predicts the [MASK]ed $\textit{(t+1)}^{th}$ token based on the bidirectional context. This process continues until a special \textit{$<$eos$>$} symbol is predicted. As with the baseline model, we again use the same beam size 5 during our evaluations.

\section{Results}
\label{sec:results}
We evaluate our models using both quantitative and qualitative approaches. In the following, we present each of them separately. 

\subsection{Quantitative Analysis}
\label{ssec:quanteval}
To evaluate our proposed models i.e., baseline and the fusion approaches, we use the standard metrics used for image captioning such as BLEU-\{1-4\}~\cite{bleu-papineni:2002}, METEOR~\cite{meteor-banerjee:2005}, ROUGE-L~\cite{rouge-lin:2004}, CIDEr~\cite{cider-vedantam:2015}, and SPICE~\cite{spice-anderson:2016}. Table~\ref{table:results} presents the average scores over three runs on the test sets of ``Karpathy split"\footnote{\url{https://cs.stanford.edu/people/karpathy/deepimagesent}\label{fnote:karpathy-split}} on the respective datasets.

\begin{table*}[t]
    \setlength{\tabcolsep}{0.7em}
    \centering
    \caption{\label{table:results} Results of proposed fusion methods on three benchmark image captioning datasets. BL-Baseline, SF-Simple fusion, CF - Cold fusion, HF - HNSG fusion, B-n - BLEU, M - METEOR, R-L - ROUGE-L, C - CIDEr, and S - SPICE.}
    \vspace*{0.5em}
    \begin{tabular}{l | c | c c c c c c c c}
    \hline  % \toprule
        \rowcolor{copper!35}
            &\multicolumn{1}{c|}{\multirow{2}{*}{}} &\multicolumn{8}{c}{Automatic Evaluation Measures} \\\cline{3-10}
        \rowcolor{copper!35}
        \multicolumn{1}{c|}{\multirow{-2}{*}{Dataset}} &\multicolumn{1}{c|}{\multirow{-2}{*}{Model}}  &B-1 & B-2   & B-3  & B-4   &M &R-L    &C  &S  \\
        \hline \addlinespace[0.3em] %\midrule
                                    &BL     &62.8   &44.9   &31.4   &21.3   &20.6   &47.1   &55.1   &14.3\\\cline{2-10}
                                    &SF     &\textbf{64.6}   &46.6   &\textbf{32.8}   &\textbf{22.8}   &21.2   &\textbf{47.8}   &\textbf{56.9}   &\textbf{14.8}\\%\cline{2-10}
        \multirow{-2}{*}{Flickr8k}  &CF     &64.5   &\textbf{46.7}   &32.7   &\textbf{22.8}   &\textbf{21.3}   &47.6   &56.5   &14.6\\%\cline{2-10}
                                    &HF     &64.1   &45.8   &32     &21.8   &20.9   &47     &55.5   &14.4  \\
        \midrule
                                    &BL     &63.3   &44.4   &30.9   &21.6   &19.2   &44.5   &45.1   &13.2  \\\cline{2-10}
                                    &SF     &\textbf{64.7}   &45.6   &\textbf{32.0}   &\textbf{22.4}   &19.7   &44.9   &\textbf{46.7}   &13.6   \\
        \multirow{-2}{*}{Flickr30k} &CF     &64.5   &\textbf{45.7}   &31.8   &22.1   &\textbf{19.8}   &\textbf{45}     &46.3   &\textbf{13.7}  \\
                                    &HF     &64.6   &45.4   &31.7   &22     &19.4   &\textbf{45}     &46.2   &13.3  \\
        \midrule
                                    &BL     &70.1   &52.8   &38.4   &28     &24.5   &51.8   &91.5   &17.5  \\\cline{2-10}
                                    &SF     &70.8   &53.7   &40.4   &30.2   &25.1   &52.6   &94.6   &17.8  \\
        \multirow{-2}{*}{MSCOCO}    &CF     &\textbf{71}     &\textbf{53.9}   &\textbf{40.7}   &\textbf{30.5}   &\textbf{25.3}   &\textbf{52.9}   &\textbf{95}     &\textbf{17.9}  \\
                                    &HF     &70.9   &53.8   &40.6   &30.5   &25     &52.7   &94.8   &17.8  \\
    \bottomrule
    \end{tabular}
\end{table*}

It can be observed from Table~\ref{table:results} that all our fusion models outperform the baseline model. However, when we compare performance of fusion models with one another we comprehend that there is no considerable difference. To be specific, on the MS-COCO dataset the Cold Fusion strategy outperforms other fusion techniques in all metrics while there is no clear winner for both Flickr8k and Flickr30k. Nevertheless we observe for the Flickr8k and Flickr30k datasets that the Simple Fusion model is a preferable option, since it gives the largest BLEU-4 score. This can be attributed to our optimization criterion since all our models are optimized for BLEU-4 while training. This leads to the increase of the BLEU-4 score; especially for the Simple Fusion model trained on Flickr8k and Flickr30k which are much smaller datasets in comparison to MS-COCO.

\subsection{Qualitative Analysis}
\label{ssec:quali}

\begin{figure}[t]
     \centering
     \begin{subfigure}[b]{0.33\textwidth}
         \centering
         \includegraphics[width=\textwidth]{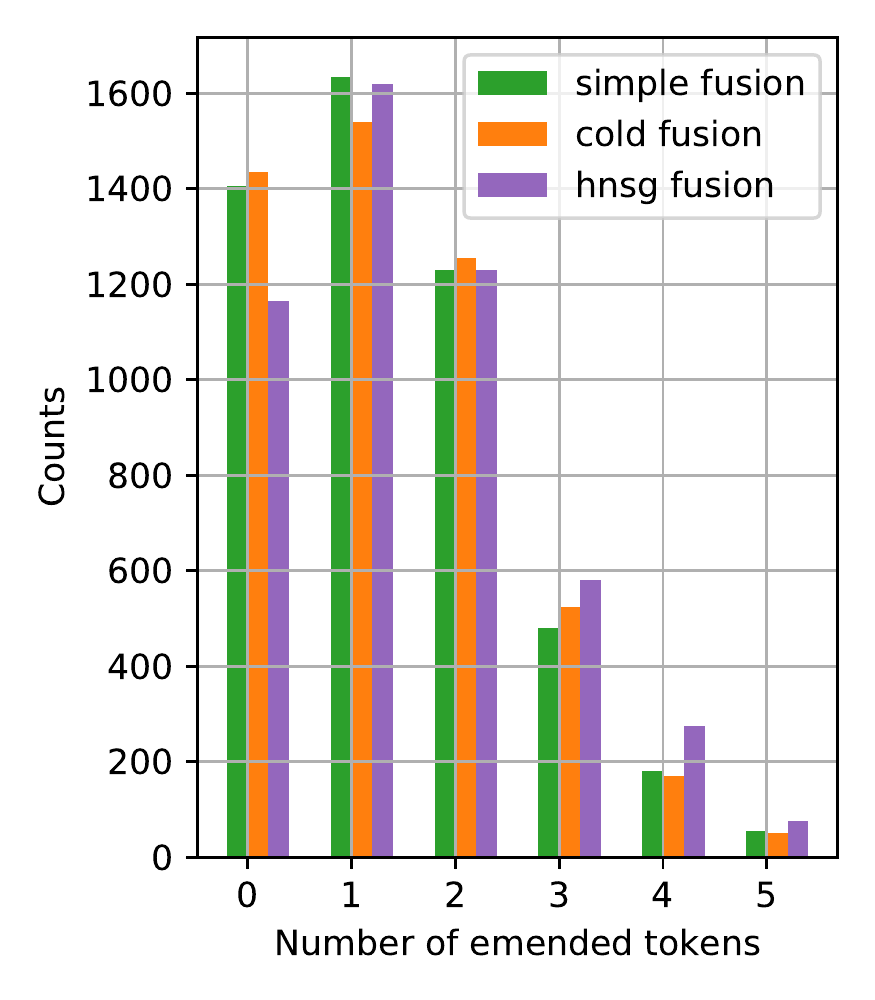}
         \caption{Flickr8k}
         \label{fig:corrections-flickr8k}
     \end{subfigure}
     \hfill
     \begin{subfigure}[b]{0.33\textwidth}
         \centering
         \includegraphics[width=\textwidth]{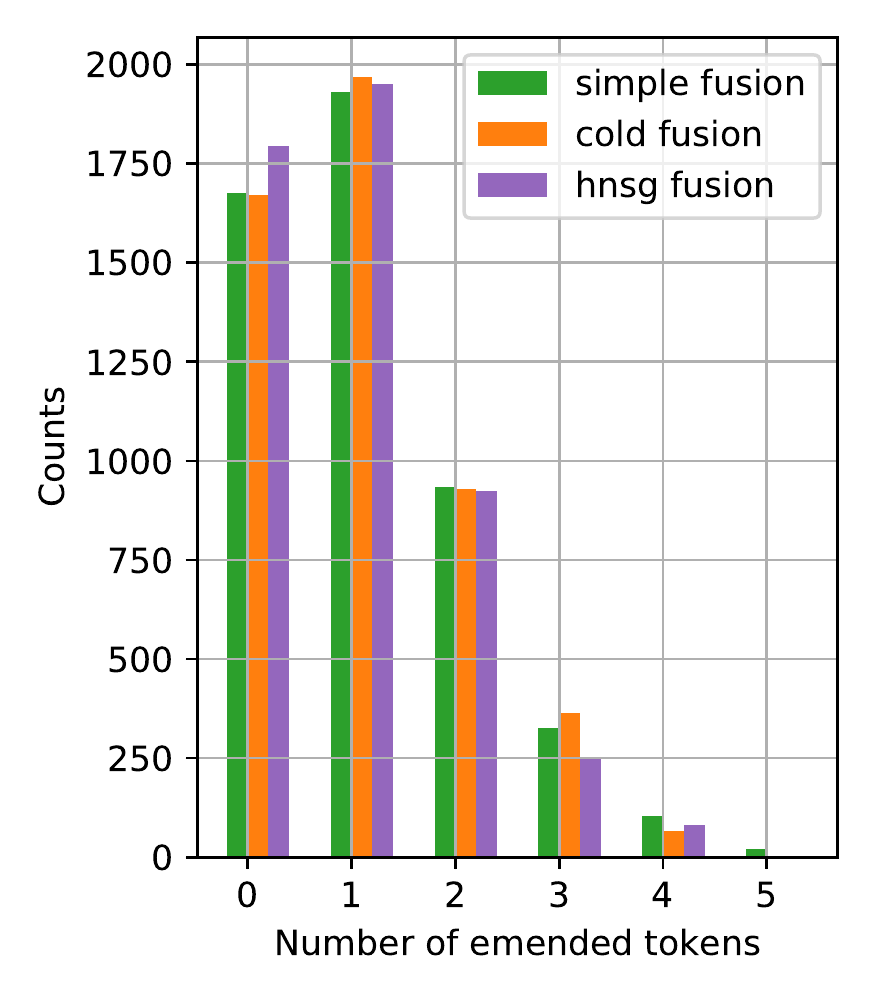}
         \caption{Flickr30k}
         \label{fig:corrections-flickr30k}
     \end{subfigure}
     \hfill
     \begin{subfigure}[b]{0.32\textwidth}
         \centering
         \includegraphics[width=\textwidth]{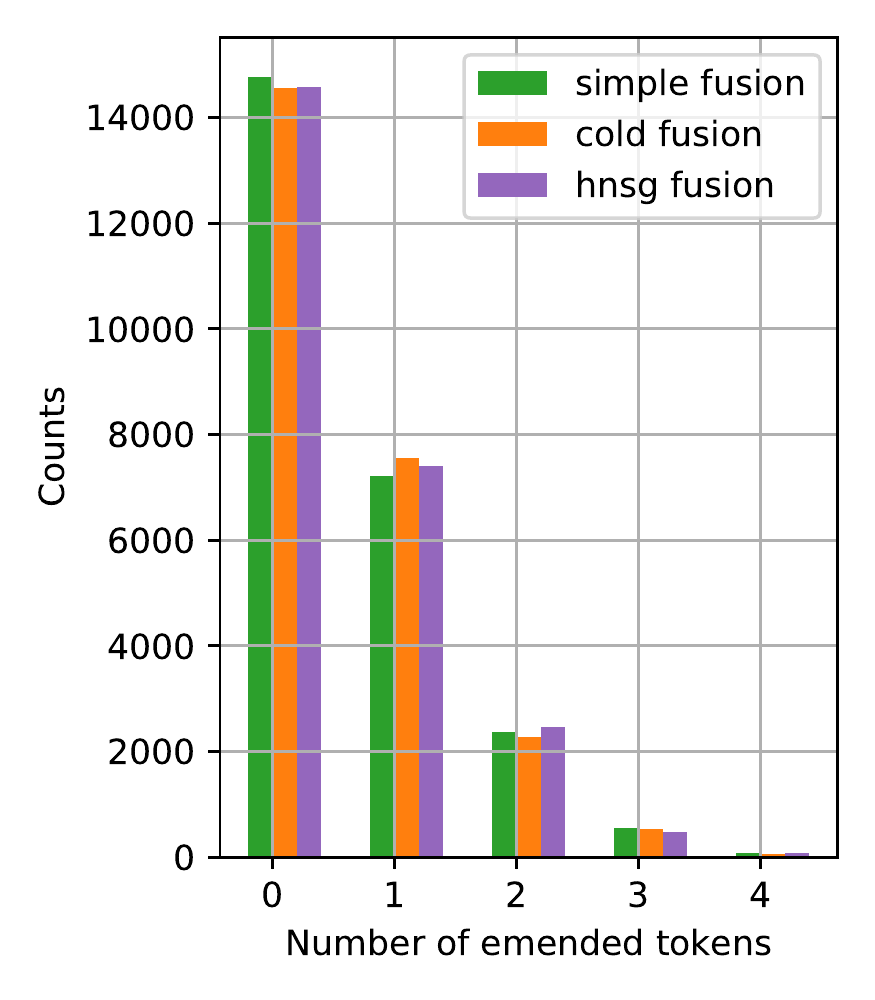}
         \caption{MSCOCO}
         \label{fig:corrections-mscoco}
     \end{subfigure}
        \caption{Distribution of (token) corrections made by fusion models on Flickr8k, Flickr30k, and MS-COCO. X-axis represents how many tokens have been changed by the fusion model while the Y-axis shows the frequencies.}
        \label{fig:corrections-distributions}
\end{figure}

In Figure~\ref{fig:corrections-distributions}, we present the token emendation distributions of the fusion techniques on all three datasets. When comparing the edits made by different fusion techniques, the distribution is similar. To understand edit distributions among datasets, we define \textit{token edit range} as the range between smallest possible token edits, which is 1, to largest possible token edits, which is the maximum length of captions. We observe that the token edit range (1-3 for MS-COCO) is smaller than Flickr8k (1-5) and Flickr30k (1-4) even though MS-COCO is about 14x and 4x larger than Flickr8k and Flickr30k respectively. This indicates the challenging nature of the Flickr caption datasets where the baseline model makes more mistakes, for which case our fusion model editing has been more helpful.

Owing to the criticism of the BLEU metric to correlate poorly with human judgements~\cite{bleu-critic-callison-burch:2006}, we perform a preliminary study on the emendations of our fusion models to better understand the \textit{quality} of emended captions. We identify several types of emendations and group them broadly into the following five categories based on whether the langauge or image-related attributes have been changed in the caption.
\begin{enumerate}
    \item Gender: Modification of gender to correctly describe image. 
    \item Color: Modification of color to correctly describe image.
    \item Specificity: Emendations to achieve specific captions instead of generic ones.
    \item Syntactic: Emendation to achieve syntactic correctness.
    \item Semantic: Emendations to correctly describe the scene.
\end{enumerate}

It should however be noted that this classification has been done with a preliminary study and a comprehensive human evaluation is needed to arrive at a more fine-grained classification. For illustrations, see Appendix~\ref{sec:append}.

\section{Conclusion}
\label{sec:conc}
In this paper, we have proposed a generic multimodal model fusion framework that can be utilized for both caption generation and editing tasks depending on the type of AuxLM that is integrated in the fusion model. We have implemented a caption editing model by integrating a pretrained BERT model and showed improved results over the baseline model on three image captioning benchmark datasets. Further, we conducted a preliminary qualitative analysis on the emended captions and identified a litany of categories based on the image or language-related attributes modified in the captions. For the future work, we plan to focus on three aspects. First, we will focus on utilizing the proposed fusion model for the \textit{caption generation} task using a state-of-the-art autoregressive language model. Second, we aspire to employ our fusion model for automatic description generation of medical images while training a domain-adapted AuxLM. Third, we plan to conduct a human evaluation on the emended captions and come up with a fine-grained classification of errors corrected by our fusion model.

\section*{Acknowledgements}
This work was funded by the Deutsche Forschungsgemeinschaft (DFG, German Research Foundation) – project-id 232722074 – SFB 1102. We extend our thanks to Matthew Kuhn for painstakingly proofing the whole manuscript.

\appendix
\section{Appendix}
\label{sec:append}
Here we provide examples of (token) corrections made by the fusion models and categorize the edits into one of the following five categories: (i) Gender (ii) Color (iii) Specificity (iv) Syntactic (v) Semantic, based on the nature of change. \newline

\noindent The above classification has been provided only for the purpose of preliminary illustration. For a thorough understanding of the trends in caption emendations and to draw conclusions, a detailed study using human evaluation should be performed on all three datasets. We leave this aspiration for future work. In the following examples, we color the incorrect tokens in \textcolor{red}{red}, the correct replacements in \textcolor{cadmiumgreen}{green}, and the equally valid tokens in \textcolor{brown}{brown}.

\subsubsection{Semantic correction.} This section presents examples where the fusion models have corrected few tokens in the baseline captions so as to make them semantically valid with respect to the image. Edits to achieve semantic correctness may include emendation of visual attributes such as colors, objects, object size, etc.

\begin{figure*}[htbp]
	\begin{subfigure}{.41\textwidth}
		\includegraphics[height=0.5\textheight,width=0.947\textwidth,keepaspectratio=true]{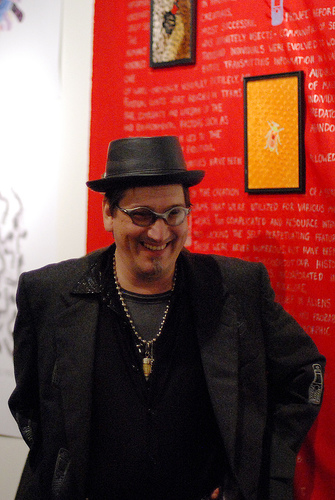} \label{fig:semantic-color-correction-2}
	\end{subfigure}
	\begin{subfigure}{.59\textwidth}
		\begin{tcolorbox}[text width=5.5cm,boxsep=2pt,left=2pt,right=2pt,top=1pt,bottom=1pt]
			\textbf{\textit{baseline}}:\\ a man \textcolor{brown}{wearing} a black \textcolor{red}{hat} and \textcolor{red}{red} hat stands in front of a brick wall \\ \\
			\textbf{\textit{simple fusion}}:\\ a man \textcolor{brown}{in} a black \textcolor{cadmiumgreen}{jacket} and \textcolor{cadmiumgreen}{black} hat stands in front of a brick wall
		\end{tcolorbox}
	\end{subfigure}
	\caption{An example illustrating the correction of semantic errors in the captions by our simple fusion model.}
\end{figure*}

\newpage
\begin{figure*}[htbp]
	\begin{subfigure}{.45\textwidth}
		\includegraphics[width=\textwidth,keepaspectratio=true]{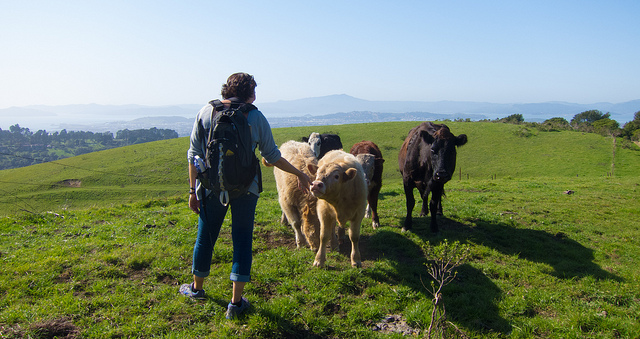} \label{fig:semantic-correction-1}
	\end{subfigure}
	\begin{subfigure}{.55\textwidth}
		\begin{tcolorbox}[text width=5.5cm,boxsep=2pt,left=2pt,right=2pt,top=1pt,bottom=1pt]
			\textbf{\textit{baseline}}:\\ a man standing next to \textcolor{red}{a sheep} in a field \\ \\
			\textbf{\textit{cold fusion}}:\\ a man standing next to \textcolor{cadmiumgreen}{cows} in a field
		\end{tcolorbox}
	\end{subfigure}
	\vspace*{-1em}
	\caption{Another example to show correction of semantic errors with cold fusion.}
\end{figure*}

\subsubsection{Gender alteration.} This section provides an illustration of the case where the fusion models corrected the wrong gender of captions from the baseline model. 

\begin{figure*}[htbp]
	\begin{subfigure}{.45\textwidth}
		\includegraphics[width=\textwidth,keepaspectratio=true]{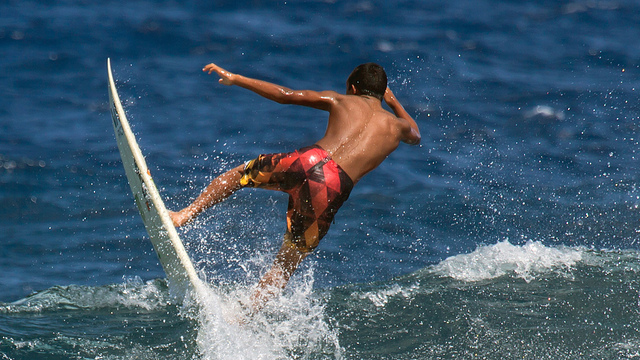}\label{fig:gender-correction-1}
	\end{subfigure}
	\begin{subfigure}{.55\textwidth}
		\begin{tcolorbox}[text width=5.5cm,boxsep=2pt,left=2pt,right=2pt,top=1pt,bottom=1pt]
			\textbf{\textit{baseline}}:\\a \textcolor{red}{woman} riding a wave on top of a surfboard \\ \\
			\textbf{\textit{cold fusion}}:\\a \textcolor{cadmiumgreen}{man} riding a wave on top of a surfboard
		\end{tcolorbox}
	\end{subfigure}
	\caption{An example of cold fusion approach achieving gender correction.}
\end{figure*}

\subsubsection{Specificity.} This deals with emendations of fusion models where the corrected captions end up describing the images more precisely than the baseline captions.

\begin{figure*}[htbp]
	\begin{subfigure}{.36\textwidth}
		\includegraphics[width=\textwidth,keepaspectratio=true]{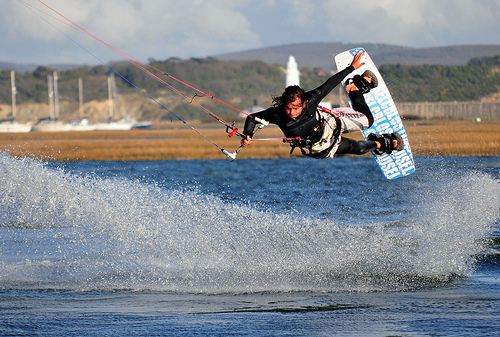} \label{fig:spec-correction-1}
	\end{subfigure}
	\begin{subfigure}{.5\textwidth}
		\begin{tcolorbox}[text width=6cm,boxsep=2pt,left=2pt,right=2pt,top=1pt,bottom=1pt]
			\textbf{\textit{baseline}}:~\\a \textcolor{brown}{person in} a \textcolor{red}{helmet} is riding a wave \\ \\
			\textbf{\textit{hierarchical fusion}}:\\a \textcolor{brown}{man} \textcolor{brown}{wearing} a \textcolor{cadmiumgreen}{harness} is riding a wave
		\end{tcolorbox}
	\end{subfigure}
	\vspace*{-1em}
	\caption{An example to show achievement of specificity with hierarchical fusion.}
\end{figure*}

\newpage
\subsubsection{Syntactic correction.} In this section, we show an example to demonstrate the case where syntactic errors such as token repetitions in the baseline captions are correctly emended by the fusion models.

\begin{figure*}[htbp]
	\begin{subfigure}{.45\textwidth}
		\includegraphics[width=\textwidth,keepaspectratio=true]{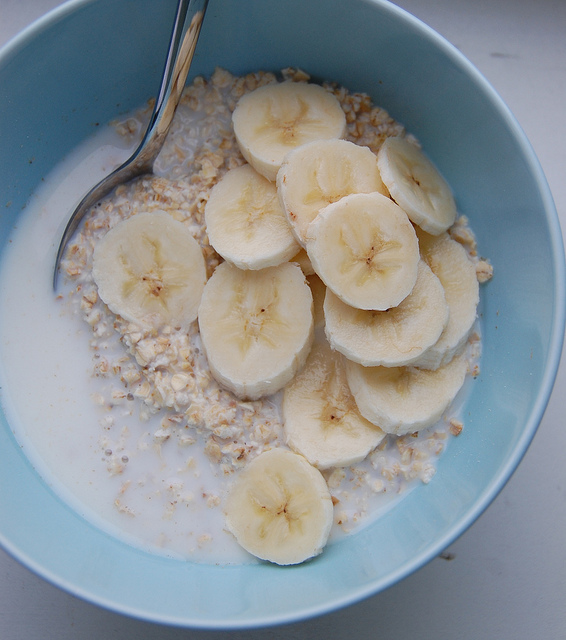} \label{fig:sync-correction-1}
	\end{subfigure}
	\begin{subfigure}{.55\textwidth}
		\begin{tcolorbox}[text width=5.9cm,boxsep=2pt,left=2pt,right=2pt,top=1pt,bottom=1pt]
			\textbf{\textit{baseline}}:\\ a white bowl filled with bananas and \textcolor{red}{bananas} \\ \\
			\textbf{\textit{simple fusion}}:\\ a white bowl filled with bananas and \textcolor{cadmiumgreen}{nuts}
		\end{tcolorbox}
	\end{subfigure}
	\vspace*{-1em}
	\caption{Replacement of repetitive tokens with a correct alternative.}
\end{figure*}

\subsubsection{Color correction.} In this part, we show an example to illustrate the case where the fusion models emended color attributes in the captions of the baseline model.

\begin{figure*}[htbp]
	\begin{subfigure}{.49\textwidth}
	    \includegraphics[width=\textwidth,keepaspectratio=true]{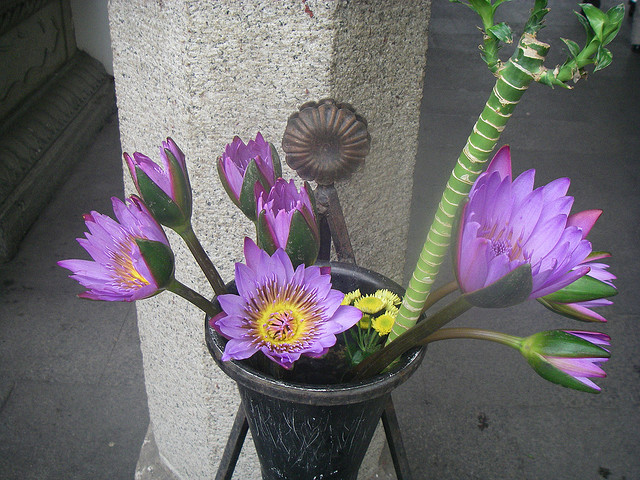} \label{fig:color-correction-1}
	\end{subfigure}
	\begin{subfigure}{.51\textwidth}
	    \begin{tcolorbox}[text width=5.5cm,boxsep=2pt,left=2pt,right=2pt,top=1pt,bottom=1pt]
	        \textbf{\textit{baseline}}:\\a vase filled with \textcolor{red}{pink} flowers on a table \\ \\
	        \textbf{\textit{cold fusion}}:\\a vase filled with \textcolor{cadmiumgreen}{purple} flowers on a table
	\end{tcolorbox}
	\end{subfigure}
	\vspace*{-1em}
	\caption{An example of cold fusion achieving emendation of color attribute.}
\end{figure*}

%%%%%%%%%%%%%%%%%%%%%
%% REFERENCES
%%%%%%%%%%%%%%%%%%%%%
\newpage
\bibliographystyle{splncs04}
\bibliography{refs}

\end{document}